\documentclass{article}

\PassOptionsToPackage{numbers}{natbib}

 \usepackage[preprint]{neurips_2026}

\usepackage[utf8]{inputenc} 
\usepackage[T1]{fontenc}    
\usepackage{hyperref}       
\usepackage{url}            
\usepackage{booktabs}       
\usepackage{amsfonts}       
\usepackage{nicefrac}       
\usepackage{microtype}      
\usepackage{xcolor}         

\usepackage{graphicx}
\usepackage{multirow}
\usepackage{amsmath}
\usepackage{makecell}
\usepackage{wrapfig}
\usepackage{enumitem}

\title{RSSA: Robust Semantic and Spatial Aligner for Collaborative Perception}

\author{%
  Zhihao Yang\textsuperscript{1} \quad
  Zhiyu Xiang\textsuperscript{1,2}\thanks{Corresponding author.} \quad
  Peng Xu\textsuperscript{1} \quad
  Tianyu Pu\textsuperscript{1} \\
  \textbf{Kai Wang}\textsuperscript{1} \quad
  \textbf{Eryun Liu}\textsuperscript{1} \quad
  \textbf{Dongping Zhang}\textsuperscript{3} \quad
  \textbf{Yong Ding}\textsuperscript{1} \\
  \textsuperscript{1}Zhejiang University \\
  \textsuperscript{2}Zhejiang Provincial Key Laboratory of Multi-Modal
  Communication Networks \\
  and Intelligent Information Processing \\
  \textsuperscript{3}China Jiliang University \\
  \texttt{yangzhihao@zju.edu.cn}, \quad
  \texttt{xiangzy@zju.edu.cn}
}

\begin{document}

\maketitle

\begin{abstract}
  V2X collaborative perception improves single-vehicle perception by aggregating multi-view features from collaborative agents. However, existing methods typically rely on simple spatial warping to align the collaborative features, which can cause semantic misalignment during fusion. Meanwhile, these deterministic warping-based methods are highly sensitive to the positional noises. In this paper, we propose a plug-in module called Robust Semantic and Spatial Aligner (RSSA), which performs semantic and probabilistic spatial transformations for robust feature alignment. Specifically, a Semantic Transformation (SemT) block modulates feature channels based on the observed relative pose to correct semantic inconsistencies. Subsequently, a Probabilistic Spatial Transformation (P-SpaT) block performs probabilistic warping by sampling from the Gaussian-modeled relative pose space. A Relative Transformation Augmentation (RT-Aug) strategy is further introduced to augment the relative poses and facilitate the training of the RSSA. Extensive experiments on both simulated and real-world datasets demonstrate that RSSA improves the state-of-the-art methods by a large margin under the configuration of both clean and noisy positions.
 
\end{abstract}

\section{Introduction}
\label{sec:intro}

\begin{wrapfigure}{r}{0.55\textwidth} 
	\centering
	\includegraphics[width=0.52\textwidth]{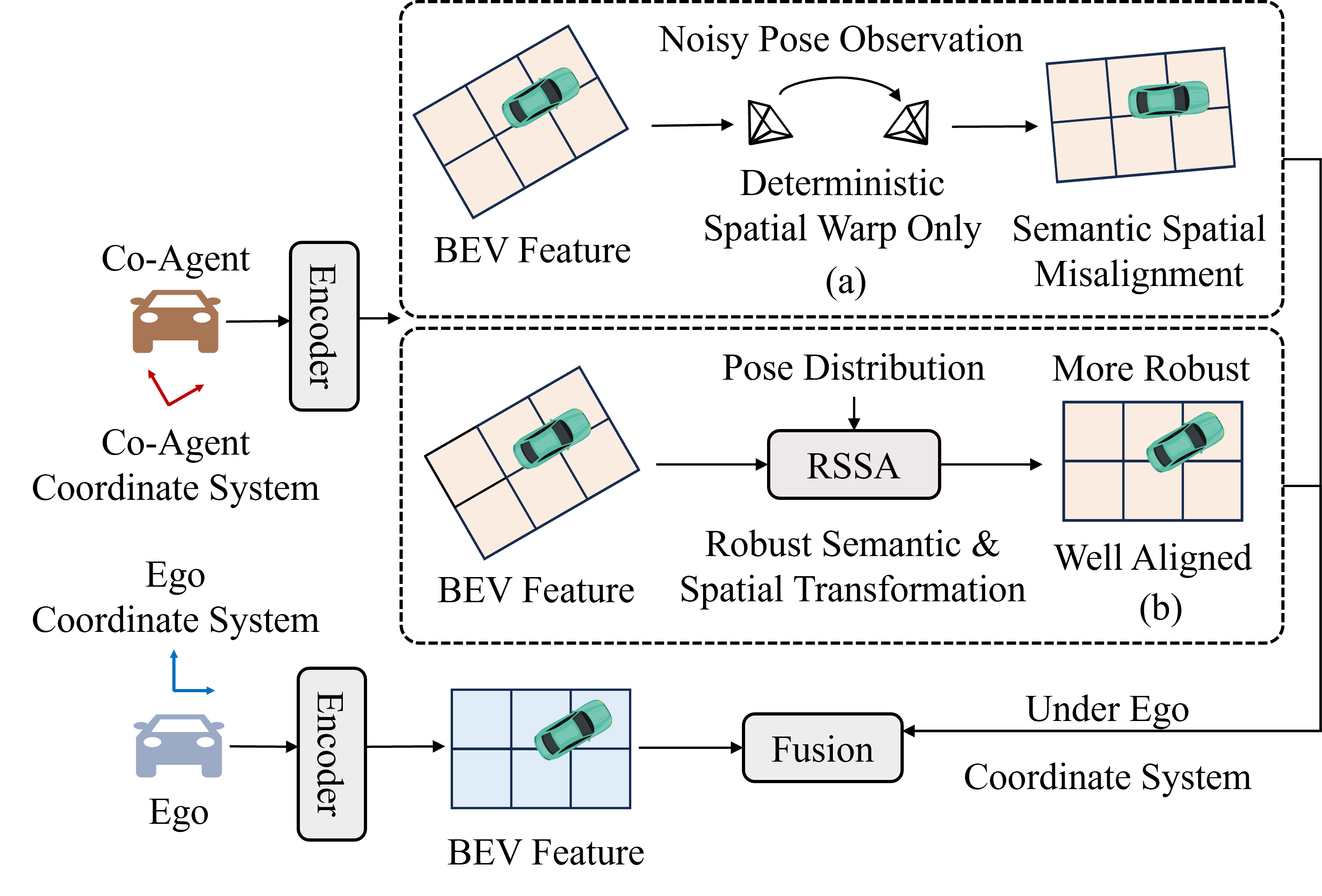}
	\caption{
			Comparison between the traditional spatial warping and our proposed RSSA for BEV features.
		}
	\label{fig:introduction}
\end{wrapfigure}

Environmental perception is critical for autonomous driving. However, single-vehicle perception~\cite{qi2017pointnet,zhou2018voxelnet,yan2018second,lang2019pointpillars,shi2019pointrcnn,shi2020pv,deng2021voxel} is inherently limited by occlusion and restricted sensing range. V2X collaborative perception effectively addresses these limitations by aggregating information from multiple collaborative agents. These complementary multi-view features enable the ego vehicle to see through occlusions and extend its detection range, significantly enhancing driving safety.

Most collaborative perception methods follow a broadcast-based paradigm~\cite{wang2020v2vnet,hu2022where2comm, xu2022cobevt,xu2025cosdh}. In this paradigm, each collaborative agent extracts local BEV features, packages them with its pose, and then broadcasts them to the ego vehicle. Upon receiving the message, the ego vehicle spatially warps the collaborative features to its own coordinate system for subsequent feature fusion.

Although efficient, this paradigm suffers from two critical limitations. First, it overlooks the semantic misalignment caused by the spatial warping. In fact, deep features such as~\cite{lang2019pointpillars} inherently encode view-dependent geometric attributes. While spatial warping successfully aligns the collaborative features' position to the ego vehicle's, the semantic representations fail to adapt, which leads to suboptimal fusion results, as illustrated in Figure~\ref{fig:introduction}(a).

Second, the above deterministic feature alignment process is sensitive to the positional noises. Existing works attempt to mitigate the impact of the noisy position through downstream processing such as consensus matching~\cite{lu2023robust,zhang2024ermvp} or robust fusion modules~\cite{xu2022v2x}, which increases the computing complexity. In fact, most modern positioning systems on-board provide location uncertainty information. Making full use of it can potentially improve the precision of feature alignment.

In this paper, we propose a plug-in module called Robust Semantic and Spatial Aligner (RSSA), which performs semantic and probabilistic spatial transformations for robust feature alignment, as illustrated in Figure~\ref{fig:introduction}(b). Specifically, RSSA incorporates a Semantic Transformation (SemT) block to correct semantic misalignment by modulating feature channels based on the observed relative pose. Subsequently, it employs a Probabilistic Spatial Transformation (P-SpaT) block to implement probabilistic warping, which samples relative poses based on the pose uncertainty. Furthermore, a Relative Transformation Augmentation (RT-Aug) strategy is introduced to improve the model's robustness against complex positional variations. We conduct extensive experiments on two representative V2X datasets: the simulated dataset V2XSet~\cite{xu2022v2x} and the real-world dataset V2X-Real~\cite{xiang2024v2x}. Experimental results demonstrate that RSSA consistently boosts the performance of existing state-of-the-art methods under both clean and noisy position settings.

In summary, the main contributions are summarized as follows:
\begin{enumerate}[topsep=3.5pt,itemsep=3pt,leftmargin=20pt]
	\item We propose RSSA, a plug-in module that performs semantic and probabilistic spatial transformations for robust collaborative feature alignment. It can effectively mitigate the semantic misalignment and reduce the sensitivity to positional noises for collaborative perception. 
	\item We introduce RT-Aug, an effective strategy that augments the relative poses with a random virtual coordinate system to boost the robustness of RSSA.
	\item We conduct extensive experiments on both simulated and real-world datasets. Experimental results demonstrate that RSSA improves the state-of-the-art methods by a large margin.
\end{enumerate}

\section{Related Work}

\subsection{Paradigms in Collaborative Perception}
As categorized in SparseAlign~\cite{yuan2025sparsealign}, collaborative perception methods can be mainly divided into two paradigms: handshake and broadcast.
The handshake paradigm~\cite{xu2022v2x,zhang2024ermvp} involves a two-round communication process. The ego vehicle first sends a request message with its pose to collaborative agents. These agents then transform their local raw data (e.g., point clouds) into the ego vehicle's coordinate system before feature extraction. While this early alignment ensures all features are learned in a unified coordinate frame, it suffers from high latency caused by its two-round communication, prohibitive computational overhead and high sensitivity to pose noises in feature fusion.

In contrast, the broadcasting paradigm~\cite{wang2020v2vnet,hu2022where2comm,li2021learning,xu2022cobevt,lu2023robust,xu2025cosdh} is more efficient and widely adopted. In this scheme, each collaborative agent extracts features in its local coordinate system and broadcasts the features along with its pose. The ego vehicle then aligns the received features to its own coordinate system using a non-learnable spatial warping before fusion. While this approach is practical for its single round of communication, it suffers from semantic misalignment caused by spatial warping. That is to say, although the spatial positions of the features have been transformed, the view-dependent semantic information embedded within the features has not been effectively converted. A more realistic scenario is the existence of both semantic and spatial misalignment when positional noises are considered. In this paper, we focus on addressing these concerns by inventing semantic and probabilistic spatial transformations.

\subsection{Feature Fusion in Collaborative Perception}
A variety of intermediate fusion methods have been proposed to aggregate features from collaborative agents. AttFuse~\cite{xu2022opv2v} introduces an agent-wise attention mechanism to adaptively weight features from different collaborators. DiscoNet~\cite{li2021learning} proposes a knowledge distillation framework where an early-fusion model acts as a teacher to improve the collaborative perception performance of the student network. Transformer-based architectures, such as V2X-ViT~\cite{xu2022v2x}, have been adopted to effectively model the interactions between features from collaborative agents. CoBEVT~\cite{xu2022cobevt} further improves upon this by introducing a fused axial attention to capture both local and global spatial relationships. To improve robustness against pose noise, CoAlign~\cite{lu2023robust} designed a pose-graph optimization module before fusion. ERMVP~\cite{zhang2024ermvp} enhances communication efficiency through feature sampling and includes a spatial calibration module to handle localization errors. CoSDH~\cite{xu2025cosdh} presented a hybrid fusion scheme to improve robustness against low communication bandwidth.

While effective, the performance of these fusion methods is critically limited by the quality of the upstream feature alignment. In this paper, we propose RSSA, a plug-in module designed for seamless integration with existing fusion networks to boost the collaborative detection accuracy.

\section{Method}
\label{method}
\subsection{Overview}
\begin{figure}[t]
	\centering
	\includegraphics[width=0.75\linewidth]{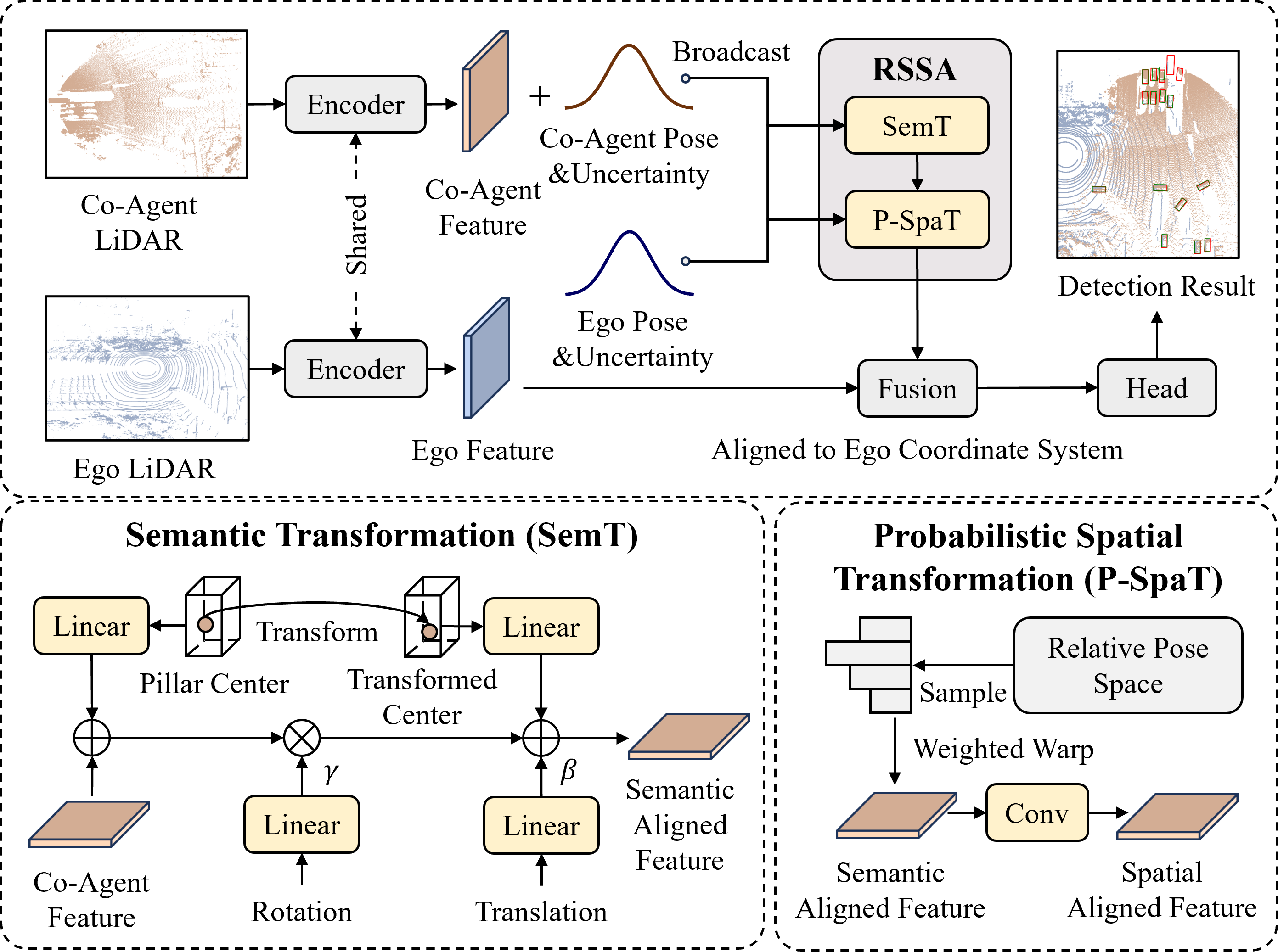}
	\caption{The overall collaborative perception pipeline and the architecture of our proposed RSSA.
		}
	\label{fig:overview}
\end{figure}

The overall collaborative perception pipeline and the architecture of RSSA are illustrated in Figure~\ref{fig:overview}. Each agent extracts a Bird's-Eye-View (BEV) feature map from its local data and packages it with the pose and the uncertainty for broadcasting.
Upon receiving the message, the ego vehicle employs the Robust Semantic and Spatial Aligner (RSSA) module to align the collaborative feature. The RSSA module consists of two sequential blocks. First, a Semantic Transformation (SemT) block corrects the semantic misalignment according to the observed poses. Subsequently, a Probabilistic Spatial Transformation (P-SpaT) block performs a probabilistic spatial warping guided by the pose uncertainty.
The aligned feature is then fed into the fusion module to generate the fused feature. Finally, a detection head is used to obtain the perception results.

\subsection{Semantic Transformation}
The Semantic Transformation (SemT) block actually performs a semantic pre-alignment prior to the spatial transformation. The components of this block is illustrated in Figure~\ref{fig:overview}.

\noindent\textbf{Transformation Encoding.} To incorporate the crucial spatial information, we need to encode the received relative poses. Considering the 6-DoF poses ${p}_i, {p}_j \in \mathbb{R}^{6}$ for the ego vehicle $i$ and collaborative agent $j$, we first parameterize them into the corresponding pose matrices $P_i, P_j \in SE(3)$ and then calculate the relative transformation matrix $\Gamma_{j\to i}$:
\begin{equation}
	\Gamma_{j\to i} = [R_{j\to i}, T_{j\to i}] = P_i \cdot P_j^{-1},
\end{equation}
where $R_{j\to i}$ and $T_{j\to i}$ represent the rotation matrix and translation vector. These components are then passed through linear layers to obtain the modulation vectors $\gamma$ and $\beta$:
\begin{align}
	\gamma &= \text{Linear}(\text{vec}(R_{j\to i})) \\
	\beta &= \text{Linear}(\text{norm}(T_{j\to i})),
\end{align}
where vec denotes flattening the rotation matrix and the norm function scales the translation vector based on the detection range.

\noindent\textbf{Position Embedding.} To provide spatial context, position embeddings are generated for both the collaborative agent $j$ and ego vehicle $i$. Let $C_j$ denotes the coordinates of the BEV pillar centers in the collaborative agent's frame, the source position embedding $PE_{j}$ is produced by passing $C_j$ through a linear layer:
\begin{equation}
	PE_{j} = \text{Linear}(C_j).
\end{equation}
Similarly, the target position embedding $PE_{i}$ is generated by transforming $C_j$ to the ego vehicle's coordinate system with $\Gamma_{j\to i}$ and then passing it through a linear layer:
\begin{equation}
	PE_{i} = \text{Linear}(\Gamma_{j\to i} \cdot C_j).
\end{equation}

\noindent\textbf{Feature Modulation.} The feature modulation integrates the position embeddings and the transformation encodings to semantically pre-align the collaborative agent's feature $F_j$ to the ego vehicle's coordinate system. First, the source position embedding $PE_j$ is added to $F_j$ for integrating initial spatial context:
\begin{equation}
	F'_{j} = F_j + PE_j.
\end{equation}

The modulation vectors $\gamma$ and $\beta$ are then used to modulate $F'_j$:
\begin{equation}
	F^{mod}_{j\to i} = \gamma \cdot F'_j + \beta.
\end{equation}

Finally, the target position embedding $PE_i$ is added to obtain the semantically aligned feature:
\begin{equation}
	F^{sem}_{j\to i} = F^{mod}_{j\to i} + PE_i,
\end{equation}

Through this transformation-aware feature modulation, the SemT block effectively pre-align the semantic information for the subsequent spatial shift of the feature.

\subsection{Probabilistic Spatial Transformation}
The co-agent's features are required to transform into the ego's frame before fusion. Existing methods simply regard the relative pose as deterministic quantities and apply it in a deterministic space transformation, which results in high sensitivity to positional noise. In this paper we explore the value of position uncertainty which can be readily obtained from the common localization systems, and encode this information into spatial transformation. We model the pose as Gaussian distribution and perform probabilistic warping by sampling from the relative pose space, as shown in Figure~\ref{fig:overview}.

\noindent\textbf{Relative Pose Space Construction.} Sampling directly from the high dimensional joint 6DoF pose space is inefficient. To reduce the complexity, we leverage the structural decomposition of the relative transformation matrix $\Gamma_{j\to i}$:
\begin{equation}
	\Gamma_{j \to i} = \begin{bmatrix} R_i^T R_j & R_i^T (t_j - t_i) \\ 0 & 1 \end{bmatrix},
\end{equation}
where $R_i, t_i$ and $R_j, t_j$ denote the rotation and translations for the ego vehicle $i$ and collaborative agent $j$, respectively. Since $\Gamma_{j\to i}$ is determined by: the orientations of both agents and their relative position difference, we construct a compact relative pose space $\mathcal{S}$ as:

\begin{equation}
	\mathcal{S} = \{ r_i, r_j, \Delta t \},  ~\text{with} ~
	\begin{aligned}
		r_i &\sim \mathcal{N}(r_i^{obs}, \sigma_{r, i}^2), \\
		r_j &\sim \mathcal{N}(r_j^{obs}, \sigma_{r, j}^2), \\
		\Delta t &\sim \mathcal{N}(\Delta t^{obs}, \sigma_{t, i}^2 + \sigma_{t, j}^2).
	\end{aligned}
\end{equation}
where $\mathcal{N}(,)$ is Gaussian distribution, $r_i^{obs}$ and $r_j^{obs}$ denote the observed rotation components of the poses, $\Delta t^{obs}$ denotes the observed relative translation, and $\sigma_{r, i}, \sigma_{r, j}$ and $\sigma_{t, i}, \sigma_{t, j}$ represent the corresponding rotation and translation uncertainties for the ego vehicle $i$ and collaborative agent $j$, respectively.

\noindent\textbf{Probabilistic Warping.} To approximate the distribution efficiently, we employ Quasi-Monte Carlo~\cite{caflisch1998monte} sampling to generate $K$ pose samples from the relative pose space $\mathcal{S}$, which include $K-1$ samples and the mean observation itself.

The spatially-aligned feature $F_{j\to i}^{spa}$ is then obtained by weighted aggregation of the features warped by these pose samples, followed by a convolutional layer:
\begin{align}
	F_{j\to i}^{spa} &= \text{Conv}\left(\sum_{k=1}^{K} w_k \cdot \text{Warp}(F_{j\to i}^{sem}, \Gamma_{j\to i}^k)\right), \\
	w_k &= \frac{\exp(p_k)}{\sum_{m=1}^{K} \exp(p_m)},
\end{align}
where $\operatorname{Warp}(\cdot)$ denotes the spatial warping, and $p_k$ is the probability of the $k$-th sample. By leveraging this uncertainty-aware probabilistic spatial warping, the P-SpaT block effectively mitigates the interference of  pose noise and significantly enhances the robustness of feature alignment.

\begin{wrapfigure}{r}{0.5\linewidth} 
	\centering
	\includegraphics[width=0.98\linewidth]{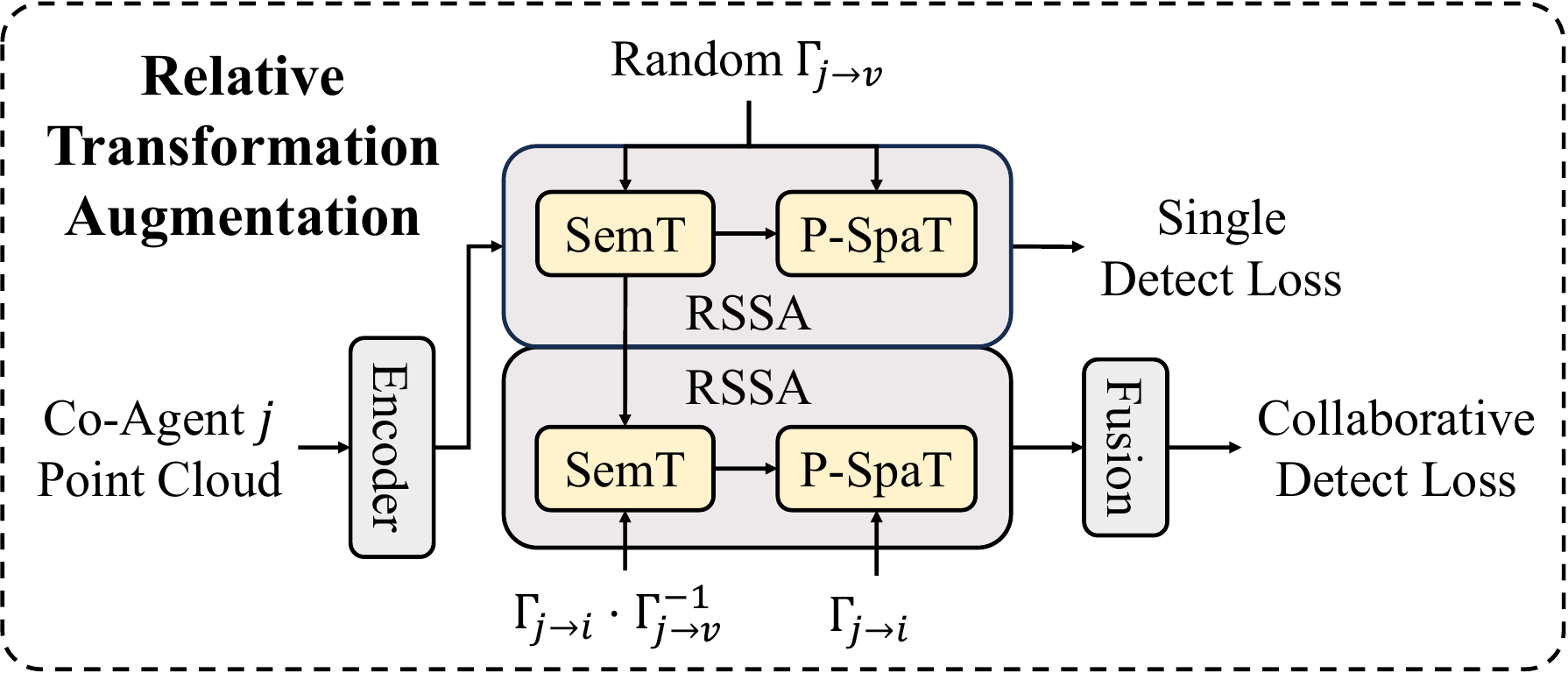}
	\caption{Illustration of the Relative Transformation Augmentation (RT-Aug) strategy.}
	\label{fig:rt_aug}
\end{wrapfigure}

\subsection{Relative Transformation Augmentation}
\label{subsec:rtaug}
In typical V2X scenarios, structured layouts result in limited relative spatial relationship, which inherently restricts the generalization ability of RSSA.
To address this geometric bias, we propose the Relative Transformation Augmentation (RT-Aug) strategy, as illustrated in Figure~\ref{fig:rt_aug}. It introduces a random virtual coordinate system $v$ to diversify the relative poses.

Specifically, RT-Aug involves a two-stage process. First, a random transformation $\Gamma_{j\to v}$ is sampled to transform the collaborative feature $F_j$ to a virtual coordinate system $v$:
\begin{align}
	F_{j\to v}^{sem} &= \text{SemT}(F_j, \Gamma_{j\to v}), \\
	F_{j\to v}^{spa} &= \text{P-SpaT}(F_{j\to v}^{sem}, \Gamma_{j\to v}).
\end{align}
$F_{j\to v}^{spa}$ is supervised by the single-agent detection loss $L_{single}$ in the virtual coordinate system to ensure the correctness of the virtual transformation.

Second, the intermediate feature $F_{j\to v}^{sem}$ is aligned to the ego vehicle's frame for subsequent feature fusion. SemT applies a compensation transformation to shift semantics from the virtual coordinate system to the ego-vehicle's, while P-SpaT applies the original relative transformation to warp the feature from the collaborative agent's coordinate system to the ego-vehicle's:
\begin{align}
	F_{j\to i}^{sem} &= \text{SemT}(F_{j\to v}^{sem}, \Gamma_{j\to i} \cdot \Gamma_{j\to v}^{-1}), \\
	F_{j\to i}^{spa} &= \text{P-SpaT}(F_{j\to i}^{sem}, \Gamma_{j\to i}).
\end{align}
Finally, $F_{j\to i}^{spa}$ is fused with the ego feature and supervised by the collaborative detection loss $L_{coll}$.

We jointly train the RSSA and the collaborative perception backbone in an end-to-end manner. The total training objective is defined as the sum of the aforementioned losses:
\begin{equation}
	L_{total} = L_{single} + L_{coll},
\end{equation}
where both $L_{single}$ and $L_{coll}$ follow the standard detection loss, employing Focal Loss~\cite{lin2017focal} for classification and Smooth L1 loss for regression.

\section{Experiment}
\label{sec:experiment}
\subsection{Experimental Setup}

\noindent\textbf{Dataset.} We evaluate our method on the public V2XSet~\cite{xu2022v2x} and V2X-Real~\cite{xiang2024v2x} datasets. V2XSet is a large-scale simulated collaborative dataset involving vehicles and infrastructure, with a detection range of $x\in[-140.8m,140.8m], y\in[-38.4m,38.4m], z\in[-5m,3m]$. V2X-Real is a large-scale real-world collaborative dataset featuring multiple vehicles and infrastructures, with a detection range of $x\in[-100.8m,100.8m], y\in[-38.4m,38.4m], z\in[-5m,3m]$.

\noindent\textbf{Implementation Details.}
Our framework is implemented in PyTorch~\cite{paszke2019pytorch} and trained on two NVIDIA RTX 3090 GPUs. We use the Adam\cite{kingma2014adam} optimizer with an initial learning rate of 0.001, which is decayed by a factor of 0.1 at epochs 10 and 20. The model is trained for a total of 40 epochs, and early stopping is used to select the best epoch. We employ PointPillars~\cite{lang2019pointpillars} as the 3D backbone to extract BEV features, with a pillar size of 0.4m x 0.4m. For the P-SpaT module, the number of pose samples $K$ is set to 9.

We conduct experiments under two settings: Clean Setting and Noisy Setting. In the Clean Setting, models are trained and evaluated without adding artificial pose noise. In the Noisy Setting, we add Gaussian noises to the agents' poses to simulate various levels of pose errors. Specifically, during training, the translation noise is sampled from $\mathcal{N}(0, \sigma_t^2)$ and applied to x and y coordinates, with $\sigma_t$ uniformly sampled from the range $[0.2m, 0.4m]$. For rotation, noise is sampled from $\mathcal{N}(0, \sigma_r^2)$ and applied to the yaw angle, with $\sigma_r$ uniformly sampled from $[0.2^\circ, 0.4^\circ]$. During inference, we evaluate the models with varying noise levels, with the standard deviations for translation and rotation noise range from 0 to 1.0 m/$^\circ$.

For the RT-Aug strategy, the virtual transformation is constructed by randomly sampling a yaw angle uniformly from $[-180^\circ, 180^\circ]$ and a z-axis translation uniformly from $[-0.5m, 0.5m]$. During inference, the virtual transformation is simply set to an identity transformation.

\noindent\textbf{Evaluation Metrics.} We use both BEV and 3D Average Precision (AP) at Intersection-over-Union (IoU) thresholds of 0.5 and 0.7 to evaluate the collaborative 3D object detection performance.

\begin{table}[h]
	\centering
	\small
	\setlength{\tabcolsep}{3pt}
	\caption{Performance comparison with state-of-the-art methods on the V2XSet and V2X-Real datasets under the clean setting. The best results are highlighted in \textbf{bold}. H and B denote handshake and broadcast, respectively. The values in \textcolor{red}{red} indicate the performance improvement of our RSSA.
}
	\label{tab:clean_performance}
	\begin{tabular}{@{}cccccc@{}}
		\toprule
		\multicolumn{2}{c}{\multirow{2}{*}{Method}} & \multicolumn{2}{c}{V2XSet} & \multicolumn{2}{c}{V2X-Real} \\
		\cmidrule(lr){3-4} \cmidrule(lr){5-6}
		\multicolumn{2}{c}{} & BEV AP@0.5/0.7 & 3D AP@0.5/0.7 & BEV AP@0.5/0.7 & 3D AP@0.5/0.7 \\
		\midrule
		\multirow{3}{*}{Att fuse~\cite{xu2022opv2v}} 
		& H & 93.20/85.46 & 92.07/75.73 & 74.29/58.83 & 70.35/34.92 \\
		\cmidrule(lr){2-6}
		& B & 84.47/68.42 & 81.13/55.09 & 68.79/53.24 & 61.80/26.04 \\
		& B+RSSA & \makecell[c]{\textcolor{red}{(+7.00/+15.19)} \\ \textbf{91.47/83.61}} & \makecell[c]{\textcolor{red}{(+8.58/+16.17)} \\ \textbf{89.71/71.26}} & \makecell[c]{\textcolor{red}{(+5.14/+7.12)} \\ \textbf{73.93/60.36}} & \makecell[c]{\textcolor{red}{(+4.34/+9.49)} \\ \textbf{66.14/35.53}} \\
		\midrule
		\multirow{3}{*}{CoBEVT~\cite{xu2022cobevt}} 
		& H & 93.07/85.85 & 91.98/76.11 & 76.43/61.10 & 72.31/37.39 \\
		\cmidrule(lr){2-6}
		& B & 86.80/69.05 & 82.89/55.62 & 72.85/55.71 & 65.78/31.00 \\
		& B+RSSA & \makecell[c]{\textcolor{red}{(+4.79/+14.18)} \\ \textbf{91.59/83.23}} & \makecell[c]{\textcolor{red}{(+7.17/+16.09)} \\ \textbf{90.06/71.71}} & \makecell[c]{\textcolor{red}{(+2.97/+5.78)} \\ \textbf{75.82/61.49}} & \makecell[c]{\textcolor{red}{(+2.37/+7.41)} \\ \textbf{68.15/38.41}} \\
		\midrule
		\multirow{3}{*}{ERMVP~\cite{zhang2024ermvp}} 
		& H & 93.72/86.65 & 92.81/76.63 & 75.96/59.69 & 71.72/37.08 \\
		\cmidrule(lr){2-6}
		& B & 86.61/67.89 & 82.29/53.41 & 71.38/53.80 & 64.29/29.90 \\
		& B+RSSA & \makecell[c]{\textcolor{red}{(+5.00/+15.65)} \\ \textbf{91.61/83.54}} & \makecell[c]{\textcolor{red}{(+7.73/+18.96)} \\ \textbf{90.02/72.37}} & \makecell[c]{\textcolor{red}{(+5.43/+8.92)} \\ \textbf{76.81/62.72}} & \makecell[c]{\textcolor{red}{(+5.34/+8.25)} \\ \textbf{69.63/38.15}} \\
		\midrule
		\multirow{3}{*}{CoSDH~\cite{xu2025cosdh}} 
		& H & 95.15/89.61 & 93.78/80.87 & 78.68/65.93 & 75.03/40.97 \\
		\cmidrule(lr){2-6}
		& B & 92.18/85.08 & 86.93/70.60 & 77.99/63.53 & 61.72/26.49 \\
		& B+RSSA & \makecell[c]{\textcolor{red}{(+1.31/+3.39)} \\ \textbf{93.49/88.47}} & \makecell[c]{\textcolor{red}{(+3.28/+5.97)} \\ \textbf{90.21/76.57}} & \makecell[c]{\textcolor{red}{(+3.54/+4.99)} \\ \textbf{81.53/68.52}} & \makecell[c]{\textcolor{red}{(+12.56/+14.38)} \\ \textbf{74.28/40.87}} \\
		\bottomrule
	\end{tabular}
\end{table}
\subsection{Quantitative Evaluation}
\noindent\textbf{Performance Comparison under Clean Setting.}
We evaluate the effectiveness of RSSA by incorporating it into several state-of-the-art methods.
The results under the clean settings are shown in Table~\ref{tab:clean_performance}, where the broadcast and broadcast+RSSA paradigms are the main focus of our evaluation, while the handshake is for reference.

We can observe that the handshake paradigm consistently outperforms the broadcast in this experiment. For instance, on the V2XSet dataset with CoBEVT, handshake surpasses broadcast by a large margin of 16.8\% in BEV AP@0.7. The key difference is that handshake performs alignment on raw point clouds, ensuring features' generation in a unified coordinate system with the ego's, thereby eliminating the semantic inconsistencies. 
This gap becomes even more pronounced when 3D AP is considered, where significant height differences between the infrastructure and the vehicle sensors exist. This comparison highlights the importance of feature semantic alignment, which is fundamentally overlooked by existing broadcast-based methods.

Our RSSA is designed to bridge this gap. By incorporating RSSA, the performance of the broadcast methods are substantially improved. Specifically, RSSA, with its lightweight design (only 1.3M parameters), boosts the BEV and 3D AP@0.7 of CoBEVT by 14.18\% and 16.09\% on V2XSet, and by 5.78\% and 7.41\% on V2X-Real, respectively. These results demonstrate that RSSA effectively corrects the semantic misalignment on the feature-level and can achieve comparable performance to that of the handshake paradigm. For more detailed analysis of RSSA's computational cost and latency, please refer to Appendix~\ref{app:efficiency}.

Notably, on the real-world V2X-Real dataset, RSSA-enhanced methods can even outperform the handshake paradigm. For instance, ERMVP+RSSA surpasses the handshake configuration by 3.03\% in BEV AP@0.7 and 1.07\% in 3D AP@0.7. This is because real-world datasets inherently contain localization and calibration noise, even in the clean setting. The handshake paradigm is highly sensitive to such noise when aligning raw point clouds. In contrast, RSSA adaptively learns a robust transformation to handle these real-world spatial inconsistencies.

\begin{figure}[t]
	\centering
	\includegraphics[width=0.95\linewidth]{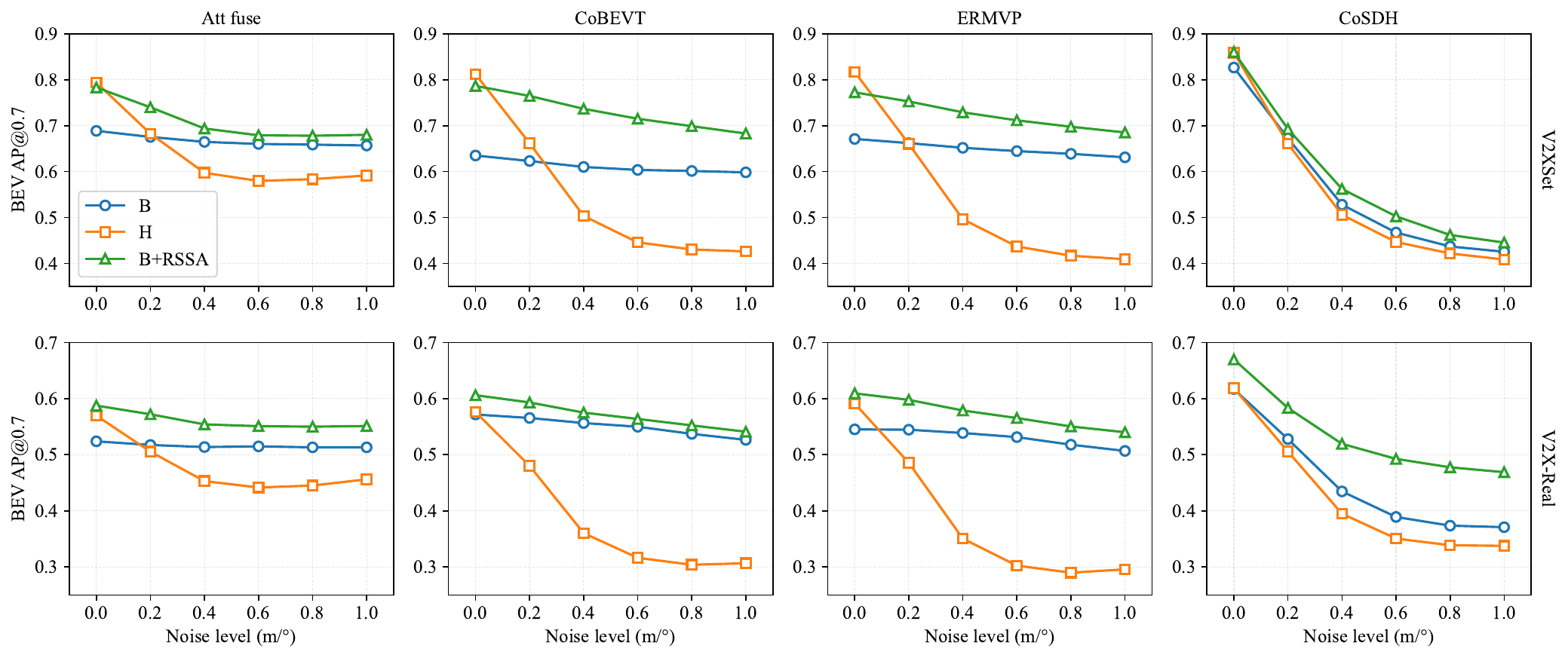}
	\caption{
		Performance comparison with state-of-the-art methods on the V2XSet (top row) and V2X-Real (bottom row) datasets under various noise levels.  Columns from left to right are: Att fuse~\cite{xu2022opv2v}, CoBEVT~\cite{xu2022cobevt}, ERMVP~\cite{zhang2024ermvp}, CoSDH~\cite{xu2025cosdh}, respectively. B and H denote Broadcast and Handshake, respectively.
	}
	\label{fig:noise_benchmark}
\end{figure}

\noindent\textbf{Performance Comparison under Noisy Setting.} We further evaluate the robustness of our method against pose errors by introducing varying levels of positional noises, as shown in Figure~\ref{fig:noise_benchmark}. As expected, the performance of the handshake paradigm drops dramatically as noise level increases. For example, on the V2XSet with CoBEVT, the BEV AP@0.7 of handshake paradigm decreases rapidly from 81.2\% to 42.6\%. In contrast, the broadcast baseline methods exhibits a slower degradation and performs better during most of the time, as it relies less on precise spatial alignment given the pre-existing semantic misalignment.

Our RSSA further outperforms the broadcast baselines across all noise levels. In low-noise settings, the significant performance gain validates the effectiveness of SemT in correcting semantic misalignment. As noise increases, RSSA maintains superior robustness thanks to the probabilistic warping of P-SpaT. Specifically, at the noise level of 1.0 m/$^\circ$ on V2X-Real, CoSDH+RSSA increases the BEV AP@0.7 by 9.8\% compared to the baseline. These results confirm the effectiveness of RSSA in addressing both semantic and spatial misalignment.

\subsection{Ablation Study}
We conduct comprehensive ablation studies on the V2XSet dataset under noisy settings, with CoBEVT as the baseline to validate the effectiveness of each proposed components.

\begin{figure}[t]
	\centering
	\begin{minipage}[t]{0.48\linewidth}
		\centering
		\includegraphics[width=\linewidth]{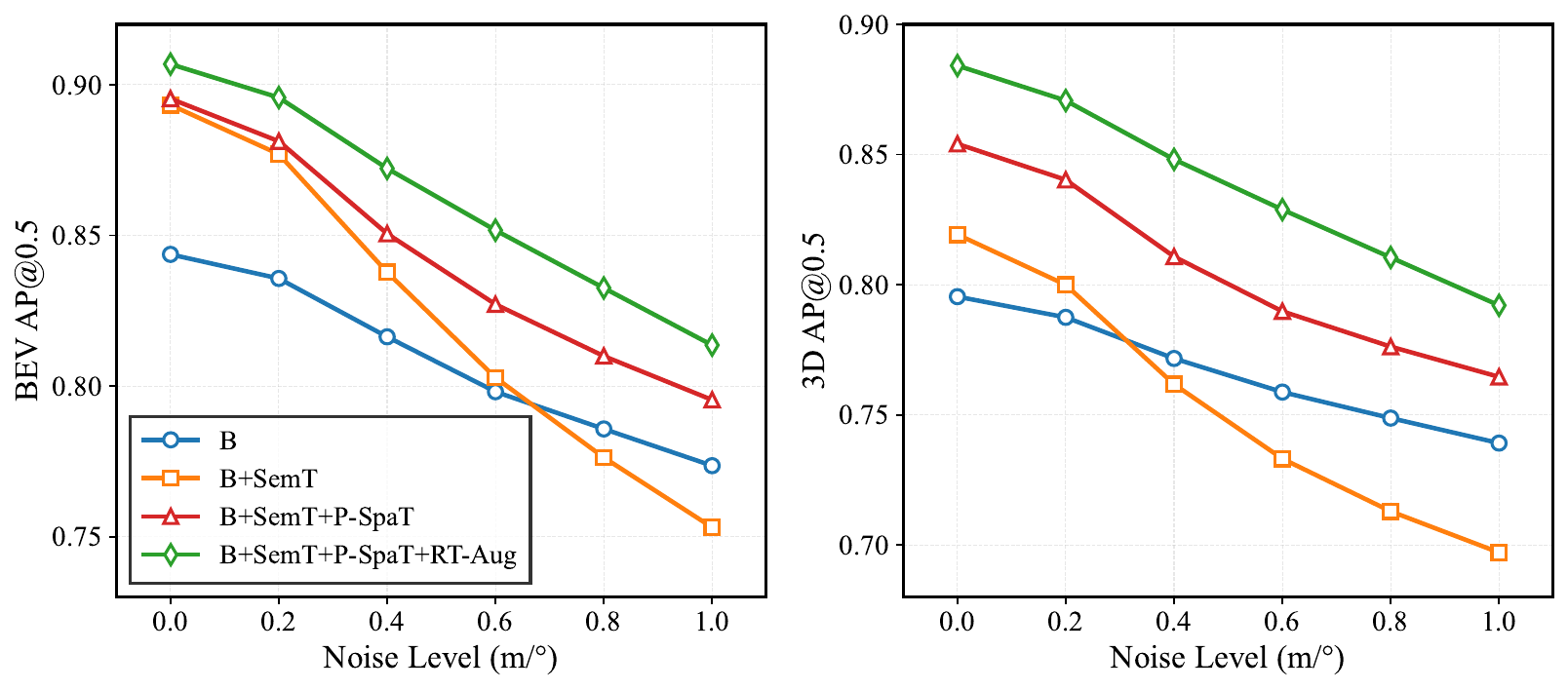}
		\caption{
			Ablation study of RSSA components. 
		}
		\label{fig:ablation}
	\end{minipage}
	\hfill 
	\begin{minipage}[t]{0.48\linewidth}
		\centering
		\includegraphics[width=\linewidth]{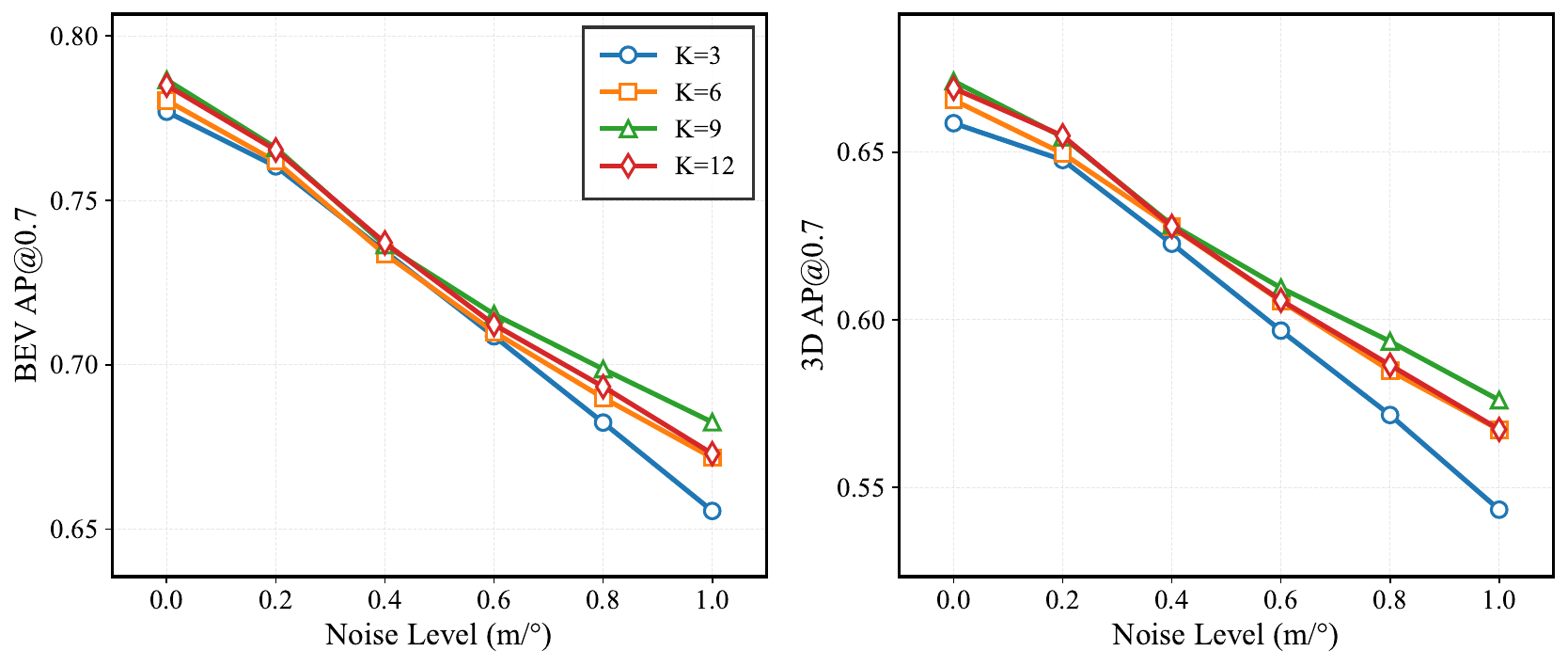}
		\caption{
			Ablation study of sampling number.
		}
		\label{fig:sample_ablation}
	\end{minipage}
\end{figure}

\noindent\textbf{Effectiveness of each component.} As shown in Figure~\ref{fig:ablation}, adding only the SemT block to the broadcast baseline results in a substantial performance boost of +4.9\% in BEV AP@0.5 and +2.4\% in 3D AP@0.5 at the zero noise level. This confirms that correcting the semantic misalignment is crucial for effective collaborative perception. However, as the SemT-only baseline still relies on a deterministic spatial warping, its performance degrades sharply as noise increases, similar to the handshake paradigm.

By incorporating the P-SpaT block, the performance curve becomes much flatter along the noise axis. At a high noise level of 1.0 m/$^\circ$, P-SpaT brings a substantial performance gain of +6.8\% in 3D AP@0.5 over the SemT-only baseline.

Finally, the inclusion of RT-Aug (the full RSSA model) yields a consistent performance improvement across all noise levels. This gain is attributed to the enhanced generalization of the SemT module. The RT-Aug strategy diversifies the training data, which guides the model to handle more complex geometric relationships, such as cases of turning maneuvers or on uneven ground.

\noindent\textbf{Sampling number in P-SpaT.} We investigate the impact of the number of pose samples $K$ in our P-SpaT module. As shown in Figure~\ref{fig:sample_ablation}, increasing the number of samples $K$ from 3 to 6 and 9 yields noticeable performance gains. Further increasing the $K$ to 12 leads to a slight performance drop. Therefore, we choose $K=9$ as it offers the best trade-off between accuracy and efficiency.

\subsection{Generalization Study}
We conduct generalization studies on the V2XSet dataset under noisy settings with CoBEVT as the baseline.

\begin{figure}[t]
	\centering
	\begin{minipage}{0.48\textwidth}
		\centering
		\includegraphics[width=\textwidth]{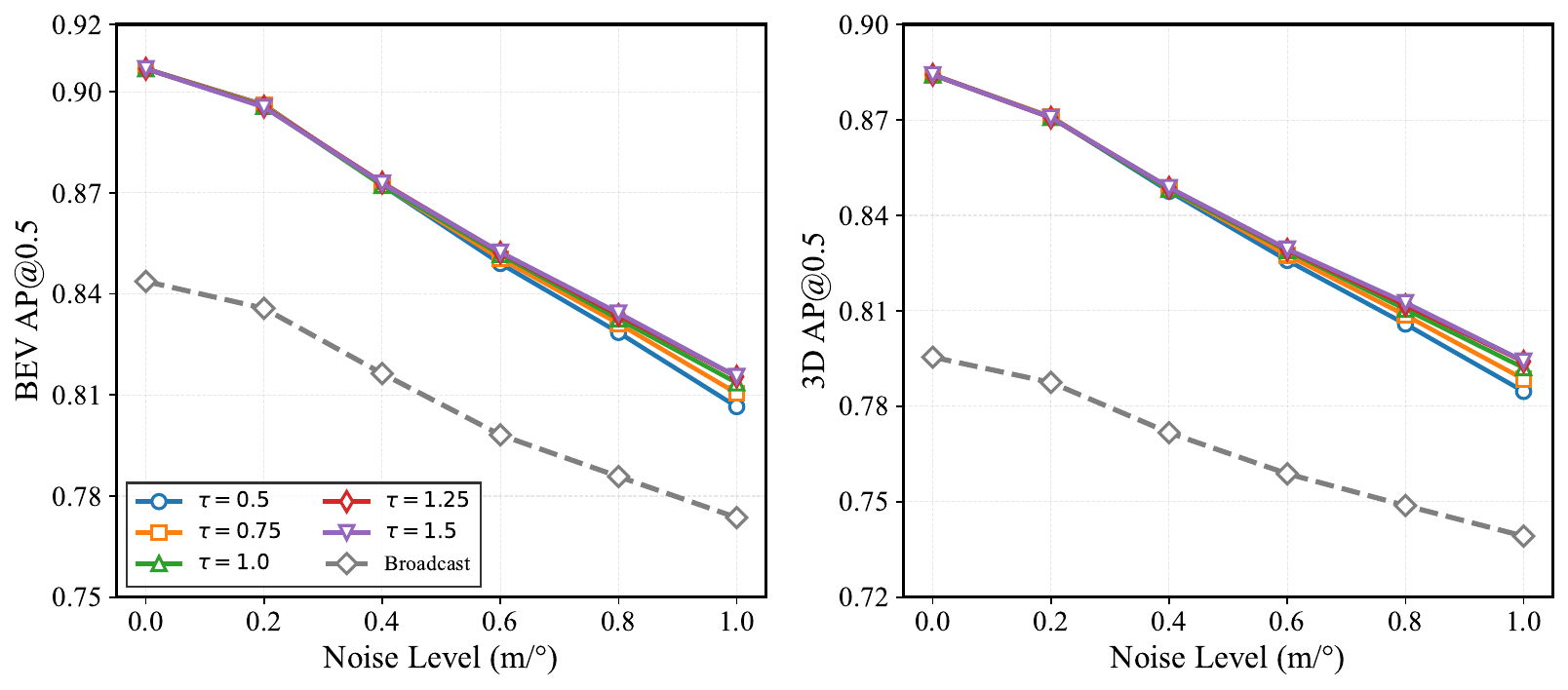}
		\caption{Generalization study of uncertainty.}
		\label{fig:inaccurate_uncertainty}
	\end{minipage}
	\hfill
	\begin{minipage}{0.48\textwidth}
		\centering
		\includegraphics[width=\textwidth]{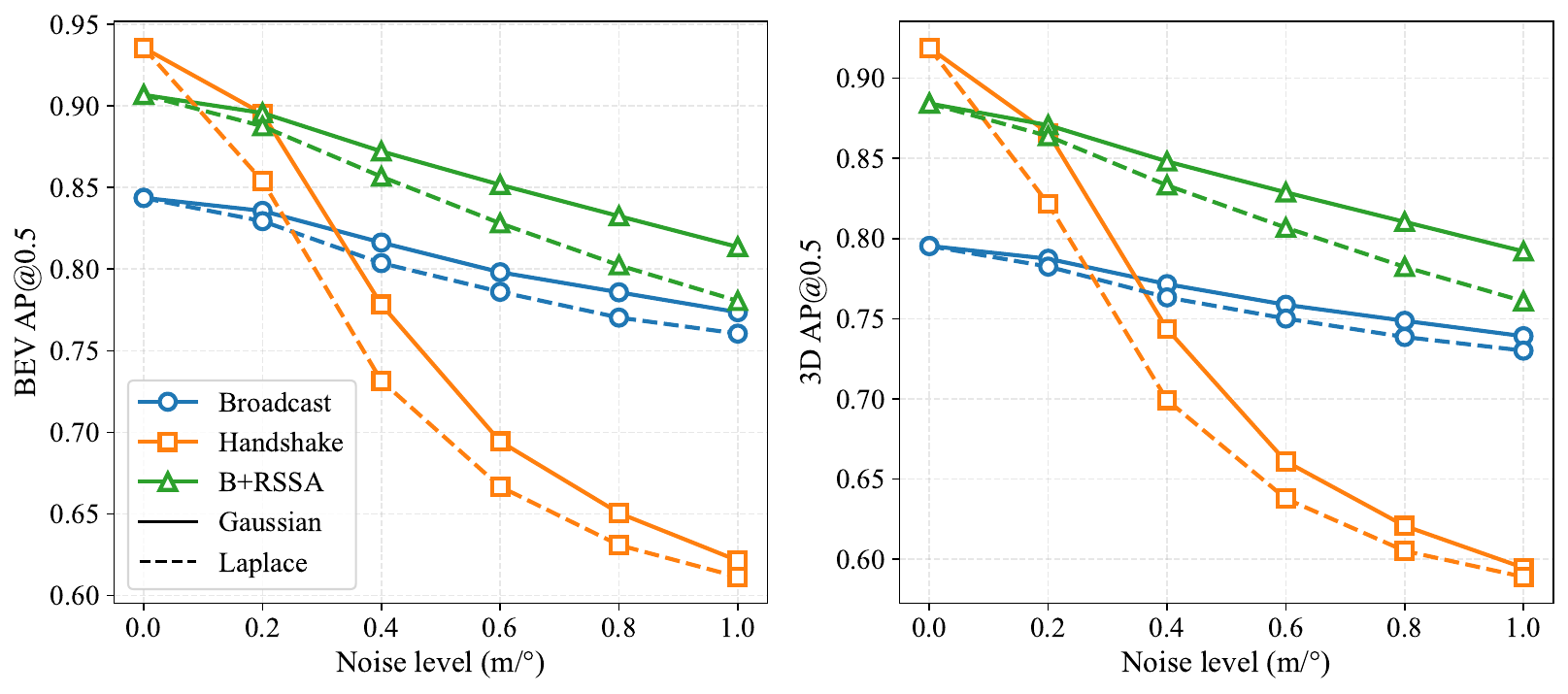}
		\caption {Generalization study of noise type.}
		\label{fig:generalization}
	\end{minipage}
\end{figure}

\textbf{Generalization to inaccurate uncertainty.}
We investigate the scenario where the provided uncertainty from the localization system is not accurate. This is simulated by multiplying the provided uncertainty with a scale coefficient $\tau\in [0.5,1.5]$. As shown in Figure~\ref{fig:inaccurate_uncertainty}, the performance slightly decreases as the uncertainty becomes inaccurate and consistently outperforms the broadcast baseline.

\textbf{Generalization to unseen noise type.}
To assess RSSA's generalization capability to different types of pose noise, we evaluate its performance with Laplace noise during inference. As shown in Figure~\ref{fig:generalization}, while all methods exhibit a performance drop when fed with the unseen Laplace noise, RSSA consistently maintains the highest performance across all noise levels.

\begin{figure}[t]
	\centering
	\includegraphics[width=0.95\linewidth]{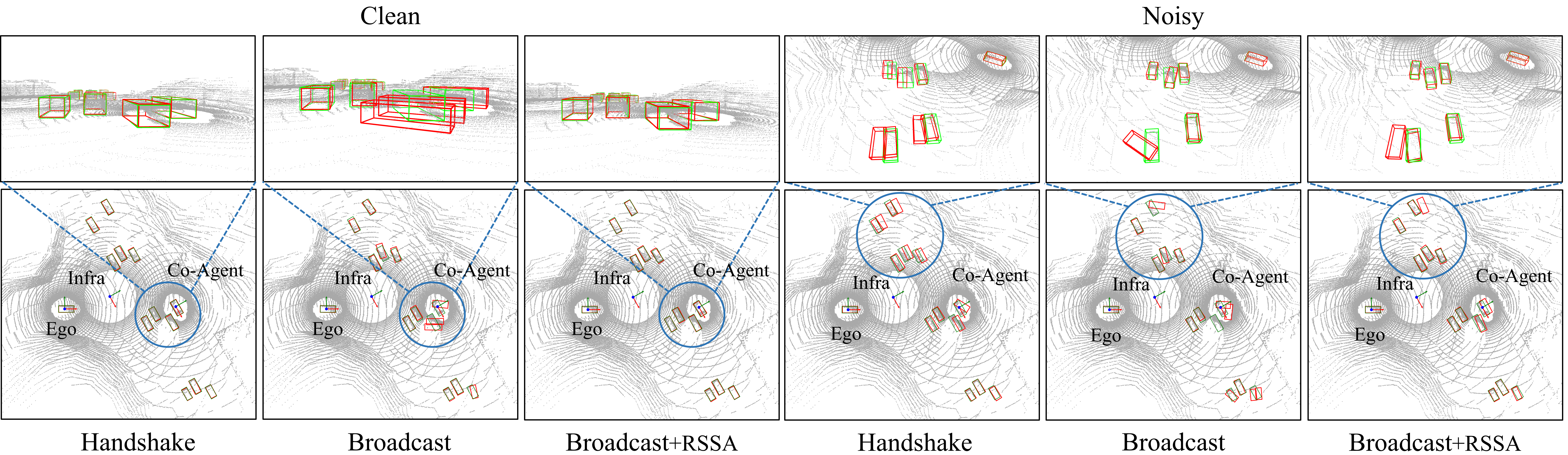}
	\caption{
		Visualization of detection results under the clean (left three columns) and noisy (right three columns) settings. 
		For each setting, the upper row shows the zoomed-in view of the region highlighted by the \textcolor{blue}{blue} circle in the lower BEV view. 
		\textcolor{green}{Green} and \textcolor{red}{red} boxes represent ground truth and predictions, respectively. 
	}
	\label{fig:visualization}
\end{figure}

\subsection{Qualitative Evaluation}
We conduct qualitative analysis on the V2XSet dataset using CoBEVT as the baseline.

\noindent\textbf{Visualization of the detection results under clean setting.} As shown in left three columns of Figure~\ref{fig:visualization}, the broadcast paradigm exhibits significant errors in orientation and height due to semantic misalignment. Specifically, for the region highlighted by the blue circle, the predicted object orientations are inconsistent with the ground truth. Similarly, in the 3D view, the object center heights are incorrectly estimated.
In contrast, our RSSA correctly aligns both the orientation in the BEV plane and the height in 3D space, producing better collaborative 3D object detection results.

\noindent\textbf{Visualization of the detection results under noisy setting.} The right three columns of Figure~\ref{fig:visualization} illustrate the results under the noise level of 1.0 m/$^\circ$. The predictions from handshake paradigm suffers from significant spatial displacements and false detections due to raw point misalignment. In contrast, our RSSA demonstrates superior robustness by effectively mitigating spatial noise and maintaining correct semantics, yielding accurate detection results that closely match the ground truth.

\section{Conclusion}
In this paper, we proposed RSSA, a novel plug-in module to strengthen the modern broadcast-based collaborative perception methods. The module is designed to address the semantic and spatial misalignment problems inherent in the deterministic spatial warping employed by most collaborative methods. RSSA integrates two key components: the Semantic Transformation (SemT) block, which corrects semantic misalignment using the observed relative pose, and the Probabilistic Spatial Transformation (P-SpaT) block, which performs robust spatial warping via probabilistic sampling guided by the pose uncertainty. The RT-Aug strategy is further proposed to augment the relative poses of the training samples. Extensive experiments on both simulated and real-world datasets demonstrate that RSSA consistently improves the performance of the state-of-the-art methods by a large margin under clean and noisy conditions.

\section*{Acknowledgments}
This work was supported by the Key Project of Natural Science Foundation of Zhejiang Province under Grant LZ26F010003, the Key Research \& Development Plan of Zhejiang Province under Grants 2024C01010 and 2024C01017, the Joint R\&D Program of the Yangtze River Delta Community of Sci-Tech Innovation under Grant 2024CSJGG01000, and the National Key Laboratory of Collective Intelligence \& Collaboration (Open Fund Project No.~QXZ25017101).

\bibliographystyle{plain}
\bibliography{references}

\newpage
\appendix

\section{Analysis of Semantic Misalignment}
\subsection{Mathematical Analysis}
\label{app:semantic_misalignment}
\begin{figure}[h]
	\centering
	\includegraphics[width=0.8\linewidth]{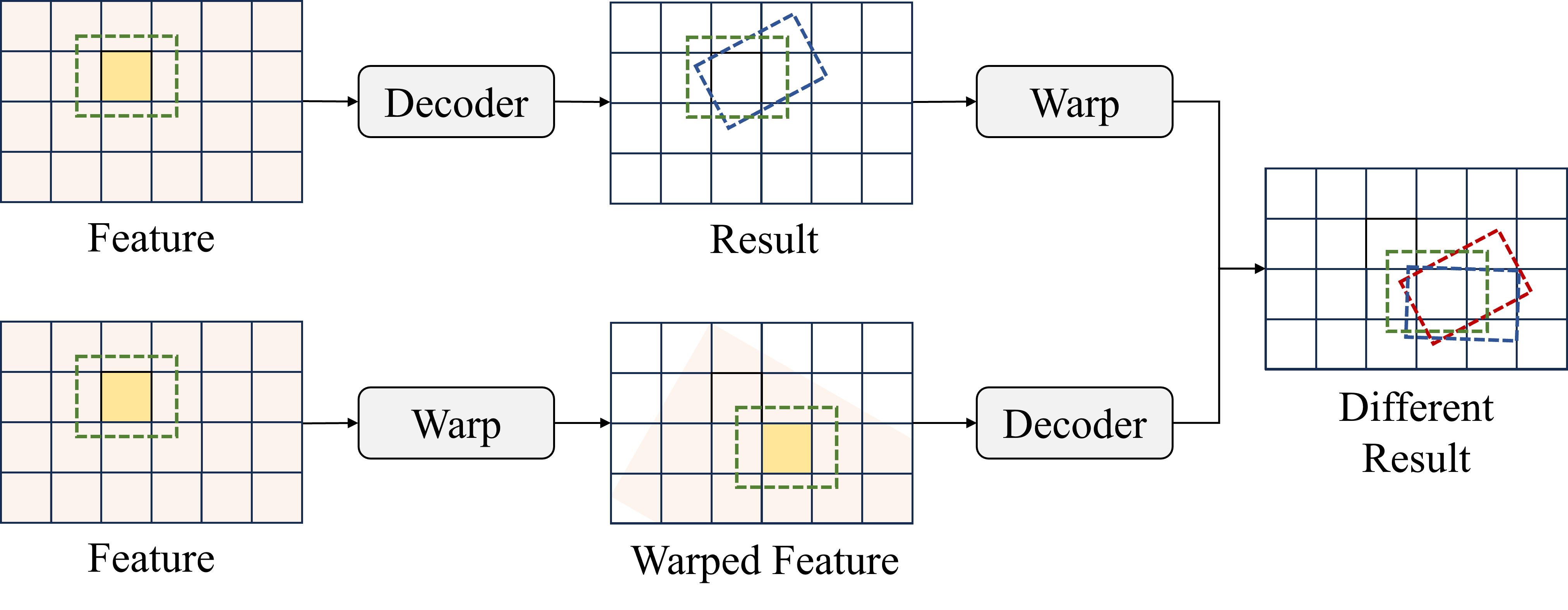}
	\caption{
	 Comparison of detection results between different align methods. The \textcolor[rgb]{0.8, 0.6, 0}{yellow square} represents the foreground feature, and the \textcolor[rgb]{0, 0.6, 0}{green box} indicates the anchor. The top row (result-level alignment) yields the target result (\textcolor{blue}{blue box}), whereas the bottom row (feature-level alignment) results in a misalignment (\textcolor{red}{red box}). The discrepancy confirms that spatial warping of BEV features causes semantic misalignment and leads to detection error, highlighting the importance and necessity of semantic alignment during coordinate transformation.
	}
	\label{fig:sup_comparison}
\end{figure}

The motivation of RSSA is that spatial warping of BEV features in the broadcast paradigm causes semantic misalignment. To demonstrate this, we compare the detection results of two align methods: \textit{result-level alignment} (correct alignment) and \textit{feature-level alignment} (spatial warping in the broadcast paradigm), as illustrated in Figure~\ref{fig:sup_comparison}.

Let $f(p)$ be the foreground local BEV feature at grid $p$ and the detection head interprets $f(p)$ using an anchor with orientation $\mathbf{o}_{anc}$. The decoded center $c$ and orientation $o$ are:
\begin{equation}
	{c} = p + \Delta p(f(p)), \quad {o} = {o}_{anc} \cdot \Delta o(f(p)).
\end{equation}
Here, $\Delta p(f(p))$ represents the predicted spatial offset from the anchor center $p$ to the object center, and $\Delta o(f(p))$ denotes the predicted orientation offset relative to the anchor's base orientation $\mathbf{o}_{anc}$.
 
For the \textit{result-level alignment}, we apply a spatial transformation ${T} = ({R}, {t})$ to the detection results:
\begin{equation}
	{c}^* = {R}(p + \Delta p(f(p))) + {t} = ({R}p + {t}) + {R} \Delta p(f(p)), \quad {o}^* = {R} \cdot ({o}_{anc} \cdot \Delta o(f(p))).
\end{equation}

For the \textit{feature-level alignment}, we apply spatial warping $\mathcal{W}$ to the feature map $f$ before decoding. Since $\mathcal{W}$ only transform the grid location $p \to p' = {R}p + {t}$ while keeping the local feature semantic $f(p)$ unchanged, the decoded results from the warped feature $\mathcal{W}(f; {T})$ at $p'$ are:
\begin{equation}
	\hat{{c}} = p' + \Delta p(f(p)) = ({R}p + {t}) + \Delta p(f(p)), \quad \hat{{o}} = {o}_{anc} \cdot \Delta o(f(p)). \label{eq:misalignment_rotation}
\end{equation}
By comparing $(\hat{{c}}, \hat{{o}})$ with the $({c}^*, {o}^*)$, we identify the problems of feature-level alignment:

\begin{enumerate}[topsep=3.5pt,itemsep=3pt,leftmargin=20pt]
	\item {Center error:} Although spatial warping successfully maps the grid position $p$ to $p' = {R}p + {t}$, the predicted spatial offset $\Delta p(f(p))$ remains anchored to the original coordinate system. Consequently, the decoded center $\hat{{c}}$ fails to incorporate the rotation ${R}$ into the predicted offset, resulting in a discrepancy of ${c}^* - \hat{{c}} = ({R} - {I}) \Delta p(f(p))\label{eq:misalignment_center}$.
	\item {Orientation error:} Similarly, while the grid position is spatially transformed to $p'$, the semantic representation $f(p)$ remains invariant, result in a discrepancy of ${o}^* \cdot \hat{{o}}^{-1} = {R}$.
\end{enumerate}

This discrepancy proves that spatial warping of BEV feature causes semantic misalignment. When ${R} \neq {I}$, this misalignment leads to substantial error in predicted object center and orientation, which demonstrate the importance and necessity of semantic alignment in collaborative perception.

\subsection{Visual Analysis}
\label{app:visualization_analysis}

\begin{figure}[h]
	\centering
	\includegraphics[width=0.8\linewidth]{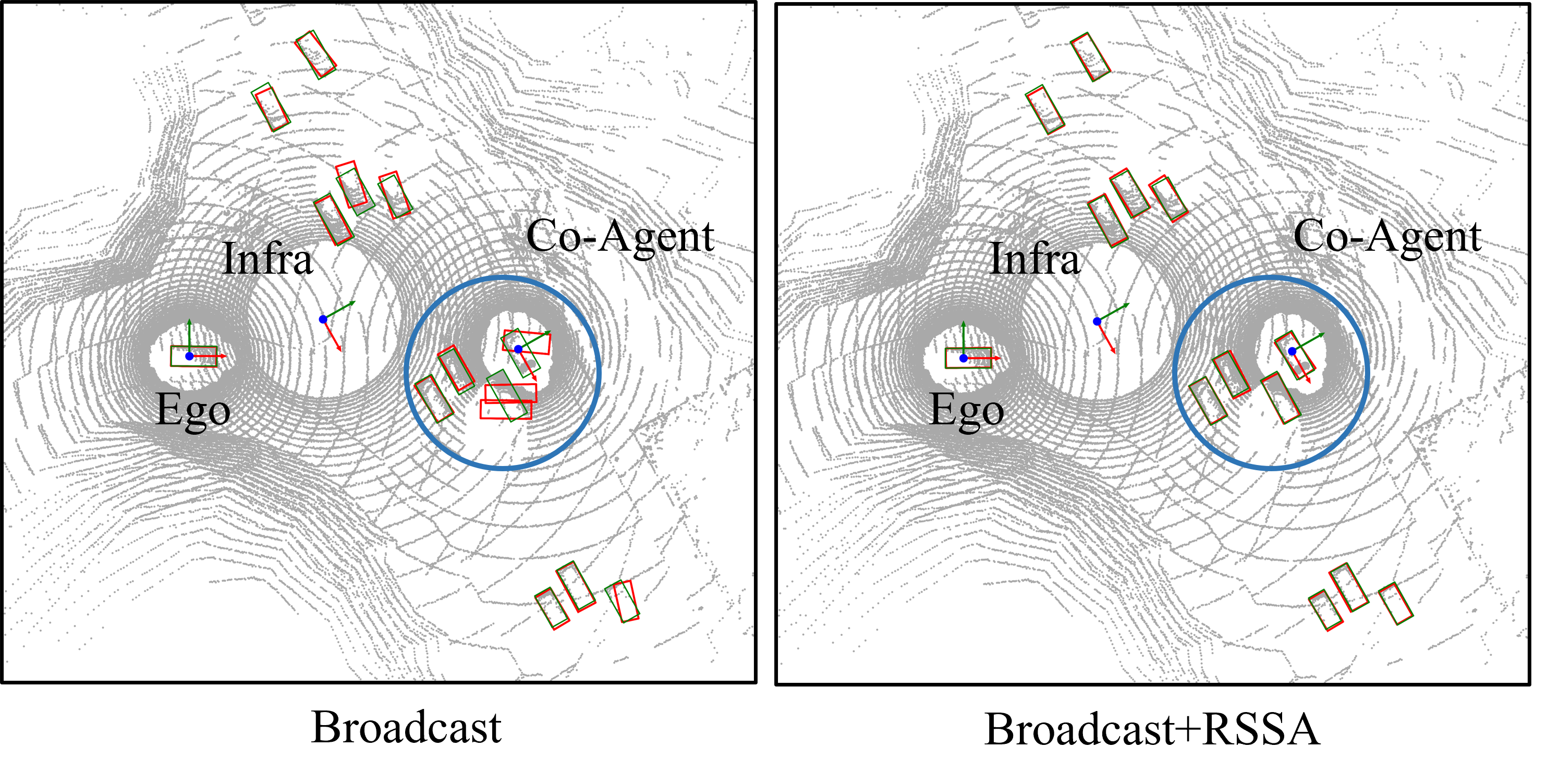}
	\caption{
		Visualization of detection results under the clean setting (zoomed-in portion of Figure \ref{fig:visualization}).
		The incorrect detection result within the \textcolor{blue}{blue circle} confirms that spatial warping causes semantic misalignment, leading to orientation and center errors.  By resolving the semantic misalignment, RSSA yields more accurate detection results.
	}
	\label{fig:sup_visualization}
\end{figure}

Figure \ref{fig:sup_visualization} provides a zoomed-in visualization of the Broadcast and Broadcast+RSSA paradigms under the clean setting, corresponding to a detailed portion of Figure \ref{fig:visualization} in the main paper.

As shown in Figure \ref{fig:sup_visualization}, the detection results of broadcast paradigm within the blue circle exhibit significant error. These objects are originally parallel to the x-axis in the coordinate systems of the co-agent and infrastructure. However, according to our proof in Eq.(\ref{eq:misalignment_rotation}), spatial warping only transforms the feature position but fails to align the feature semantic. Consequently, during fusion, the ego vehicle interprets the orientation semantic as being relative to its own coordinate system, resulting in incorrect predictions that are parallel to its own x-axis. 

Furthermore, the predicted centers also exhibit a noticeable shift from the ground truth. This is consistent with the proved center error ${c}^* - \hat{{c}} = ({R} - {I}) \Delta p(f(p))$, confirming that the prediction centers are shifted due to the lack of rotational compensation for the spatial offset. In contrast, by aligning both semantics and spatial coordinates, our RSSA effectively corrects these errors and produces more accurate detection results, as shown on the right.

\subsection{Direct Validation of SemT}
\label{app:semt_validation}

To directly validate the ability of SemT, we analyze the detection
results obtained from the features before SemT and after SemT using
Broadcast+RSSA with CoBEVT~\cite{xu2022cobevt} on
V2XSet~\cite{xu2022v2x} under the clean setting.
We match detection results with ground-truth boxes using an IoU
threshold of 0.5 and report the mean center and orientation error
over the matched results (True Positives), as shown in
Table~\ref{tab:semt_validation}.

\begin{table}[h]
  \centering
  \caption{Mean error before and after SemT on V2XSet under
    the clean setting.}
  \label{tab:semt_validation}
  \begin{tabular}{lcc}
    \toprule
    Feature & Center error (m) & Orientation error ($^\circ$) \\
    \midrule
    Before SemT & 0.52 & 45.93 \\
    After SemT  & 0.17 & 10.78 \\
    \bottomrule
  \end{tabular}
\end{table}

Even among TPs, SemT reduces the mean center error by 0.35 m and
the mean orientation error by 35.15$^\circ$. Before SemT, the detection
results exhibit a large center error and a more pronounced orientation
error, consistent with our mathematical analysis in
Appendix~\ref{app:semantic_misalignment}: spatial warping transforms
feature locations but leaves the center-offset and orientation
semantics in the source coordinate system. After SemT, both errors are
substantially reduced, providing direct evidence that SemT learns the
intended semantic transformation.

\section{Analysis of Relative Pose Distributions}
\label{app:pose_analysis}

\begin{figure}[h]
	\centering
	\includegraphics[width=1.0\linewidth]{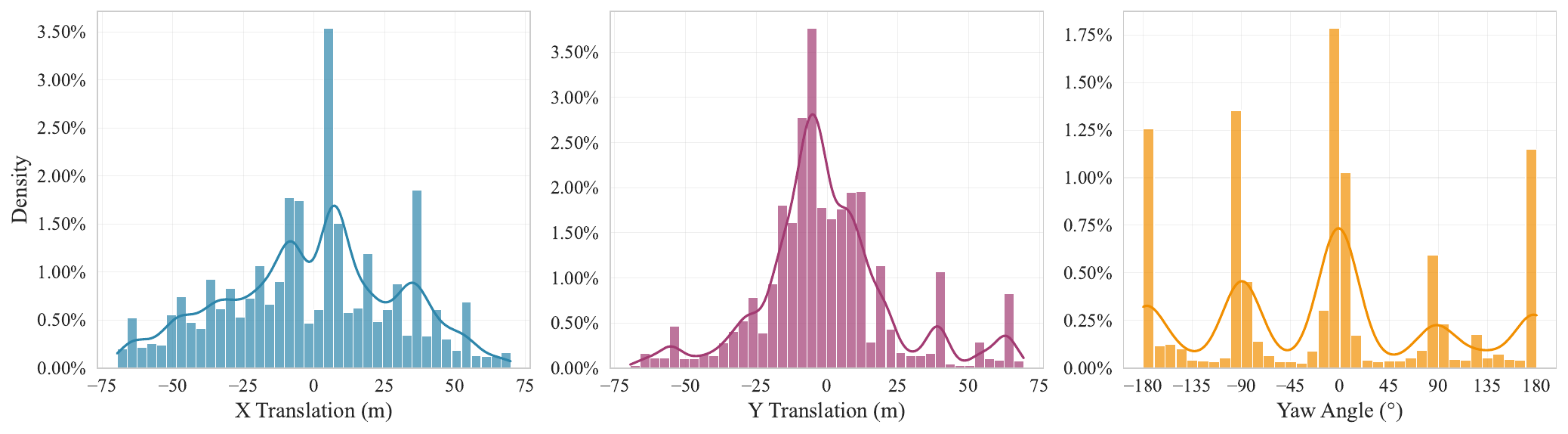}
	\caption{
		Statistical analysis of relative pose distributions on the V2XSet dataset. The yaw angle distribution (right) exhibits significant data bias with peaks at $0^\circ, \pm 90^\circ, \text{ and } \pm 180^\circ$, highlighting the structured nature of urban scenes and the necessity of the RT-Aug strategy.
	}
	\label{fig:sup_pose_distribution}
\end{figure}

As discussed in the section~\ref{subsec:rtaug}, V2X scenarios are often highly structured, which results in limited relative spatial relationships between agents. To empirically demonstrate this, we provide a statistical analysis of the relative poses between collaborative agents and the ego vehicle on the V2XSet~\cite{xu2022v2x} dataset, as illustrated in Figure \ref{fig:sup_pose_distribution}.

The distributions of relative X and Y translations (left and middle) exhibit a wide range of values due to the diverse movement of agents. More importantly, the yaw angle distribution (right) also shows a varied spread resulting from maneuvers such as turns. As mathematically proven in Appendix \ref{app:semantic_misalignment}, whenever a relative rotation exists ($R \neq I$), spatial warping alone fails to align feature semantics, which empirically justifies the necessity of RSSA for semantic alignment.

Simultaneously, the yaw distribution reveals a distinct data bias, with pronounced peaks at $0^\circ$, $\pm 90^\circ$, and $\pm 180^\circ$. This phenomenon is expected in structured urban environments, where agents typically move in parallel, opposite, or perpendicular directions at intersections. Such biased distribution can cause the network to overfit to these specific orientations, potentially hindering its ability to handle arbitrary relative angles, such as those encountered during complex turning maneuvers. To address this limitation, we proposed the RT-Aug strategy. By introducing a random virtual coordinate system during training, RT-Aug effectively diversifies the relative transformations, enabling the model to learn generalized feature alignment.

\section{Additional Experiments}

\subsection{Ablation Studies}

\noindent\textbf{Position Embeddings in SemT.} To assess the contribution of the position embeddings in SemT,
we compare RSSA with and without them on V2XSet~\cite{xu2022v2x}
using CoBEVT~\cite{xu2022cobevt} under the clean setting.

\begin{table}[h]
  \centering
  \caption{Ablation of position embeddings on V2XSet with CoBEVT
    under the clean setting.}
  \label{tab:pe_ablation}
  \begin{tabular}{lcc}
    \toprule
    Method & BEV AP@0.7 & 3D AP@0.7 \\
    \midrule
    Broadcast+RSSA & 83.23 & 71.71 \\
    Broadcast+RSSA, w/o position embeddings & 81.69 & 67.87 \\
    \bottomrule
  \end{tabular}
\end{table}

As shown in Table~\ref{tab:pe_ablation}, removing the position
embeddings within SemT decreases 3D AP@0.7 by 3.84 points.
This confirms that the position embeddings provide complementary
global spatial context for semantic transformation rather than
duplicating the relative-transformation encoding.

\noindent\textbf{Isolated Impact of P-SpaT.}
To isolate the contribution of P-SpaT, we compare RSSA with and
without P-SpaT on V2XSet using CoBEVT under the noisy setting.

\begin{table}[h]
  \centering
  \caption{Ablation of P-SpaT on V2XSet with CoBEVT
    (BEV AP@0.7). Noise levels are given in m/$^\circ$.}
  \label{tab:pspat_ablation}
  \begin{tabular}{lccccc}
    \toprule
    Method & 0.2 & 0.4 & 0.6 & 0.8 & 1.0 \\
    \midrule
    RSSA w/o P-SpaT & 72.74 & 70.52 & 68.26 & 66.33 & 64.41 \\
    RSSA w/ P-SpaT  & 76.46 & 73.66 & 71.50 & 69.88 & 68.28 \\
    \bottomrule
  \end{tabular}
\end{table}

As shown in Table~\ref{tab:pspat_ablation}, incorporating P-SpaT
consistently improves performance across all noise levels.
At the 1.0 m/$^\circ$ noise level, P-SpaT improves AP by 3.87 points,
demonstrating the effectiveness of probabilistic spatial alignment.

\noindent\textbf{Component-Level Ablation with ERMVP.}
To verify the effectiveness of our design, we conduct the same
component-level ablation on V2XSet using ERMVP~\cite{zhang2024ermvp} under the noisy setting.

\begin{table}[h]
  \centering
  \small
  \caption{Component-level ablation on V2XSet with ERMVP
    (BEV AP@0.5). Noise levels are given in m/$^\circ$.}
  \label{tab:ermvp_ablation}
  \begin{tabular}{lcccccc}
    \toprule
    Method & 0.0 & 0.2 & 0.4 & 0.6 & 0.8 & 1.0 \\
    \midrule
    Broadcast
      & 85.84 & 85.24 & 83.48 & 81.71 & 79.94 & 78.20 \\
    Broadcast+SemT
      & 89.09 & 87.48 & 82.48 & 77.75 & 74.46 & 71.75 \\
    Broadcast+SemT+P-SpaT
      & 89.65 & 88.06 & 85.03 & 82.61 & 81.05 & 79.90 \\
    Broadcast+RSSA
      & 89.84 & 88.73 & 86.49 & 84.68 & 82.92 & 81.33 \\
    \bottomrule
  \end{tabular}
\end{table}

As shown in Table~\ref{tab:ermvp_ablation}, the component-level
ablation trends with ERMVP are consistent with those observed using
CoBEVT. At zero noise, SemT improves AP by 3.25 points, validating
its contribution to semantic alignment. As noise increases, SemT
alone degrades due to deterministic spatial warping, whereas P-SpaT
substantially improves robustness, outperforming SemT by 8.15 points
at the 1.0 m/$^\circ$ noise level. Finally, incorporating RT-Aug
achieves the best performance across all noise levels, exceeding
SemT+P-SpaT by 1.43 points at 1.0 m/$^\circ$. These results confirm
the effectiveness and generalizability of the RSSA components.

\subsection{Robustness to Other Noise}

\noindent\textbf{Miscalibrated Uncertainty.}
We evaluate CoBEVT on V2XSet under zero-mean pose noise with the
actual standard deviation fixed at 0.4 m/$^\circ$. To simulate
miscalibrated uncertainty, we provide P-SpaT with input uncertainty
levels of 1.0 m/$^\circ$, 2.0 m/$^\circ$, and 4.0 m/$^\circ$.

\begin{table}[h]
  \centering
  \caption{Robustness to miscalibrated pose uncertainty
    (BEV AP@0.7).}
  \label{tab:miscalibrated_uncertainty}
  \begin{tabular}{lccc}
    \toprule
    Method & 1.0 m/$^\circ$ & 2.0 m/$^\circ$ & 4.0 m/$^\circ$ \\
    \midrule
    Handshake      & 50.32 & 50.32 & 50.32 \\
    Broadcast      & 61.00 & 61.00 & 61.00 \\
    Broadcast+RSSA & 73.93 & 73.60 & 73.25 \\
    \bottomrule
  \end{tabular}
\end{table}

As shown in Table~\ref{tab:miscalibrated_uncertainty}, although
RSSA's performance gradually decreases as the input uncertainty
becomes less accurate, it consistently outperforms both Broadcast
and Handshake. Since Broadcast and Handshake do not use the input
uncertainty, their performance remains unchanged across the three
settings. Even with an input uncertainty of 4.0 m/$^\circ$, RSSA
remains 12.25 and 22.93 AP points higher than Broadcast and
Handshake, respectively, demonstrating its robustness to
miscalibrated uncertainty.

\noindent\textbf{Biased Localization Errors.}
We further evaluate robustness to systematic localization bias
on V2XSet with CoBEVT. We introduce non-zero-mean pose noise
with the standard deviation fixed at 0.4 m/$^\circ$, while setting
the mean bias to 0.2 m/$^\circ$, 0.4 m/$^\circ$, and
0.6 m/$^\circ$.

\begin{table}[h]
  \centering
  \caption{Robustness to biased localization errors (BEV AP@0.7).}
  \label{tab:biased_localization}
  \begin{tabular}{lccc}
    \toprule
    Method & 0.2 m/$^\circ$ & 0.4 m/$^\circ$ & 0.6 m/$^\circ$ \\
    \midrule
    Handshake      & 48.86 & 45.26 & 41.75 \\
    Broadcast      & 60.88 & 60.43 & 60.06 \\
    Broadcast+RSSA & 73.13 & 72.24 & 70.78 \\
    \bottomrule
  \end{tabular}
\end{table}

As shown in Table~\ref{tab:biased_localization},
Broadcast+RSSA achieves the best performance across all bias
levels. At the largest mean bias of 0.6 m/$^\circ$, it outperforms
Broadcast and Handshake by 10.72 and 29.03 AP points, respectively,
demonstrating its robustness to biased localization errors.

\noindent\textbf{Missing Data.}
We evaluate robustness to missing data on V2XSet with CoBEVT
by randomly dropping collaborative agent's data with probabilities
of 0.25, 0.50, and 0.75.

\begin{table}[h]
  \centering
  \caption{Robustness to missing data (BEV AP@0.7).}
  \label{tab:missing_data}
  \begin{tabular}{lccc}
    \toprule
    Method & 0.25 & 0.50 & 0.75 \\
    \midrule
    Handshake      & 77.81 & 74.03 & 69.19 \\
    Broadcast      & 63.78 & 64.00 & 64.30 \\
    Broadcast+RSSA & 76.02 & 72.83 & 68.97 \\
    \bottomrule
  \end{tabular}
\end{table}

As shown in Table~\ref{tab:missing_data}, Broadcast+RSSA achieves
performance comparable to Handshake and consistently outperforms
Broadcast across all missing probabilities. At the highest missing
probability of 0.75, it remains comparable to Handshake while
outperforming Broadcast by 4.67 AP points, demonstrating its
robustness to missing data.

\noindent\textbf{Temporal Desynchronization.}
We evaluate robustness to temporal desynchronization on V2XSet
with CoBEVT by introducing time delays to collaborative agent's
data from 100 ms to 400 ms.

\begin{table}[h]
  \centering
  \caption{Robustness to temporal desynchronization (BEV AP@0.7).}
  \label{tab:temporal_desync}
  \begin{tabular}{lcccc}
    \toprule
    Method & 100 ms & 200 ms & 300 ms & 400 ms \\
    \midrule
    Handshake      & 68.52 & 56.34 & 53.04 & 52.43 \\
    Broadcast      & 61.45 & 59.57 & 59.18 & 58.88 \\
    Broadcast+RSSA & 74.74 & 69.25 & 63.95 & 60.35 \\
    \bottomrule
  \end{tabular}
\end{table}

As shown in Table~\ref{tab:temporal_desync}, the data
desynchronization caused by time delays degrades collaborative
perception performance. Nevertheless, Broadcast+RSSA achieves
the best performance across all delay levels. At the largest
delay of 400 ms, it outperforms Broadcast and Handshake by
1.47 and 7.92 AP points, respectively, demonstrating its
robustness to temporal desynchronization.

\subsection{Effect of Noisy Training}

As commonly adopted in previous works, CoAlign~\cite{lu2023robust}
and V2X-ViT~\cite{xu2022v2x}, we introduce synthetic pose noise
during training for all evaluated methods. Clean training optimizes
the model for accurate poses, whereas noisy training exposes it to
pose errors. Consequently, the two training settings can produce
different results even at zero additional test noise.

To quantify this trade-off, we evaluate CoBEVT+RSSA trained under
the clean and noisy settings across different test noise levels,
as shown in Table~\ref{tab:training_setting}.

\begin{table}[h]
  \centering
  \small
  \caption{Robustness evaluation of CoBEVT+RSSA under different
    training settings (BEV AP@0.7). Test noise levels are given
    in m/$^\circ$.}
  \label{tab:training_setting}
  \begin{tabular}{llcccccc}
    \toprule
    Dataset & Training & 0.0 & 0.2 & 0.4 & 0.6 & 0.8 & 1.0 \\
    \midrule
    V2XSet~\cite{xu2022v2x} & Clean & 83.23 & 62.75 & 41.13
           & 33.97 & 32.68 & 34.16 \\
    V2XSet & Noisy & 78.65 & 76.46 & 73.66
           & 71.50 & 69.88 & 68.28 \\
    V2X-Real~\cite{xiang2024v2x} & Clean & 61.49 & 51.75 & 41.34
             & 38.71 & 39.28 & 41.28 \\
    V2X-Real & Noisy & 60.61 & 59.32 & 57.51
             & 56.38 & 55.23 & 54.10 \\
    \bottomrule
  \end{tabular}
\end{table}

On V2XSet, noisy training sacrifices 4.58 AP points at zero noise
but improves AP by 34.12 points at the 1.0 m/$^\circ$ noise level.
On V2X-Real, it improves AP by 12.82 points at 1.0 m/$^\circ$,
with only a 0.88-point decrease at zero additional noise.
Thus, noisy training provides substantial gains under high pose
noise, with a marginal decrease at zero additional noise on V2X-Real.

Nevertheless, we agree that synthetic noise cannot fully represent
real-world pose errors. On the other side, current carefully
calibrated and constructed benchmark datasets often contain less
pose noise in their clean settings than real-world scenarios.
Therefore, noisy training remains valuable for practical deployment.

\subsection{Complementarity with CoAlign}
\label{app:coalign}

\begin{figure}[h]
	\centering
	\includegraphics[width=0.9\linewidth]{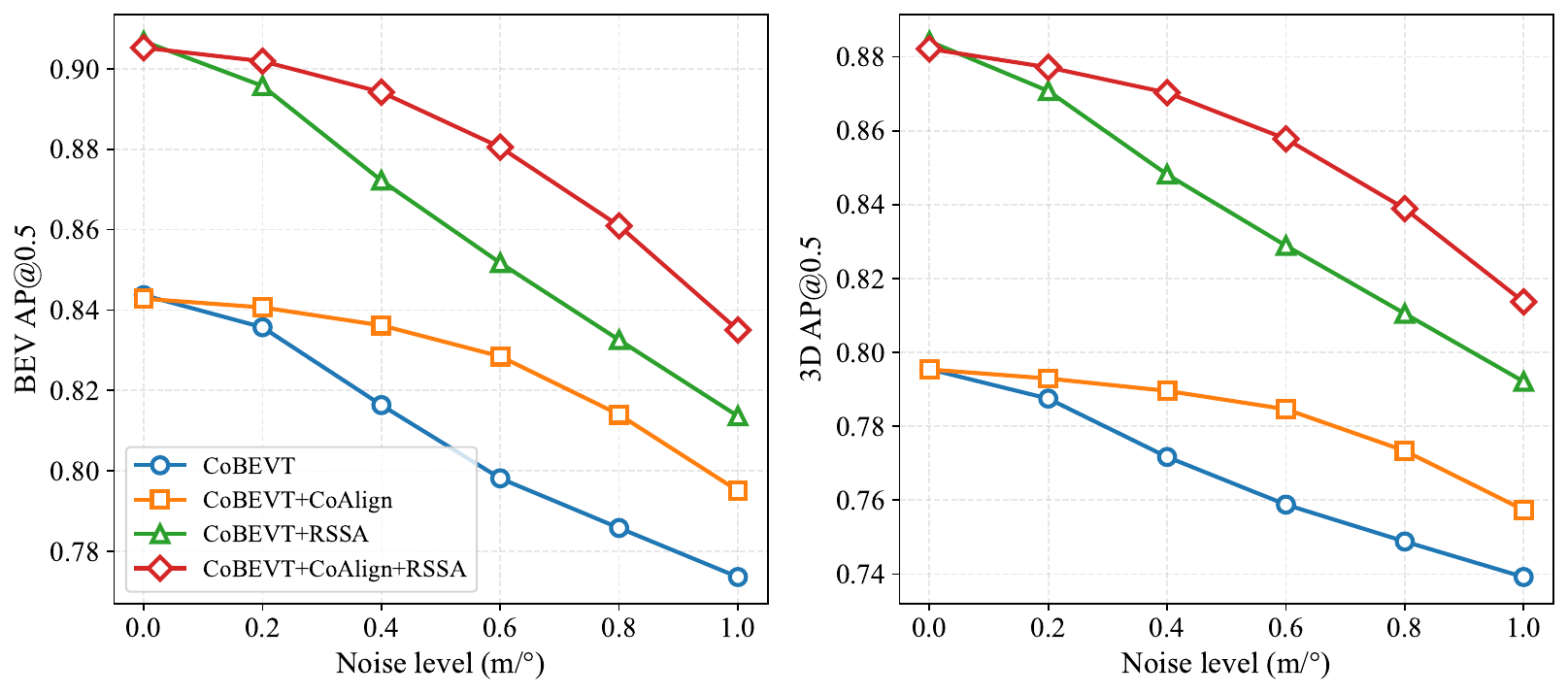}
	\caption{
		Performance comparison of CoAlign and RSSA using CoBEVT as the fusion module on the V2XSet dataset.
	}
	\label{fig:sup_coalign}
\end{figure}

To demonstrate the universality of RSSA, we evaluate its complementarity with other robustness method, specifically the pose-graph optimization in CoAlign. As illustrated in Figure \ref{fig:sup_coalign}, both CoAlign and RSSA independently provide performance gains over the baseline, and their joint integration yields the best results across all noise levels. This improvement stems from the fact that CoAlign optimizes alignment at the pose level, while RSSA ensures robust and semantically consistent alignment at the feature level. These results prove that RSSA is a general plug-in module that can effectively enhance other robustness methods.

\subsection{Model Complexity and Efficiency}
\label{app:efficiency}

\begin{table}[h]
	\centering
	\small
	\caption{Comparison of model complexity, efficiency, and performance using CoBEVT as fusion module on the V2XSet dataset.}
	\label{tab:efficiency}
	\begin{tabular}{@{}lccccc@{}}
		\toprule
		\multirow{2}{*}{Method} & Parameters & Flops & Latency & \multicolumn{2}{c}{BEV AP@0.7} \\
		\cmidrule(lr){5-6}
		& (M) & (G) & (ms) & Clean & Noisy (1.0 m/$^\circ$) \\
		\midrule
		Broadcast & 10.51 & 251.05 & 97.18 & 69.05 & 59.82 \\
		Broadcast+RSSA       & 11.84 & 281.27 & 111.76 & \textbf{83.23} & \textbf{68.28} \\
		\midrule
		RSSA & \textcolor{blue}{1.33}  & \textcolor{blue}{30.22}  & \textcolor{blue}{14.58} & \textcolor{red}{+14.18} & \textcolor{red}{+8.46} \\
		\bottomrule
	\end{tabular}
\end{table}

We analyze the complexity and efficiency of RSSA to demonstrate its practicality as a lightweight plug-in. Table \ref{tab:efficiency} compares the parameters, GFlops, latency, and detection performance with and without RSSA using CoBEVT as fusion module on the V2XSet dataset.

As shown in Table \ref{tab:efficiency}, RSSA is highly efficient. Since the module consists primarily of simple linear layers and convolutional layers, the parameter increment is only 1.33 M. Furthermore, thanks to the parallel execution of the probabilistic warping, RSSA introduces a marginal computational overhead of 30.22 GFlops and only 14.58 ms of additional latency. With this minimal cost, our module effectively achieves semantic and robust spatial alignment, obtaining substantial performance gains of +14.18\% in the clean setting and +8.46\% under high pose noise level (1.0 m/$^\circ$) on BEV AP@0.7. These results confirm that RSSA provides a superior trade-off between detection accuracy and computational efficiency.

\section{Qualitative Evaluation on V2X-Real Dataset}
\label{app:visualization_v2xreal}

\begin{figure}[h]
	\centering
	\includegraphics[width=0.9\linewidth]{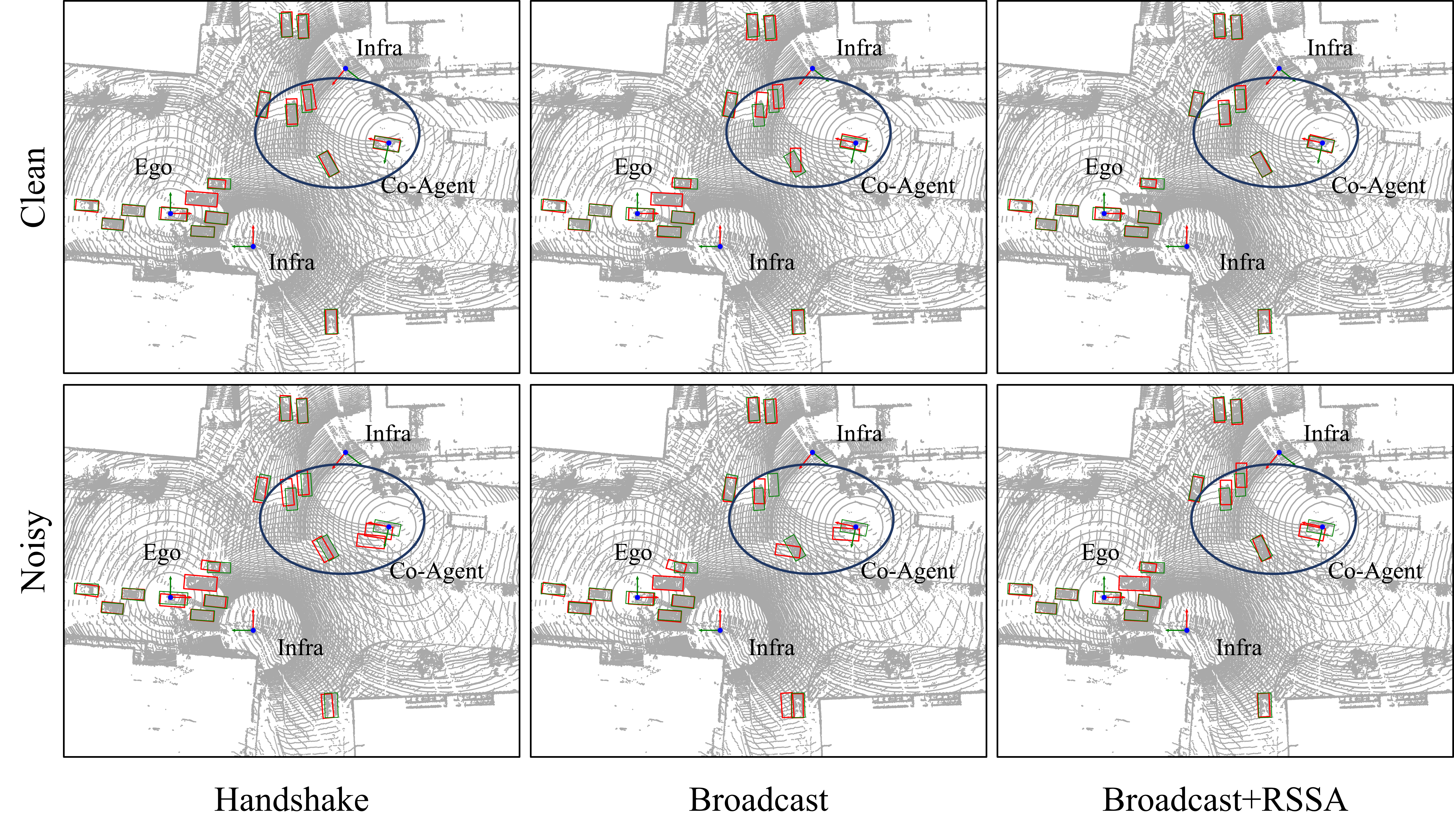}
	\caption{
		Visualization of detection results on the V2X-Real dataset using CoBEVT as fusion module across clean (top) and noisy (bottom) settings. The \textcolor{blue}{blue circles} highlight that the broadcast paradigm suffers from semantic misalignment caused by spatial warping, and the handshake paradigm is sensitive to noise, resulting in more false positives. Our RSSA provides the most robust and accurate detection by performing semantic transformation and probabilistic spatial transformation.
	}
	\label{fig:sup_vis_bev}
\end{figure}

We provide additional visualization results on the real-world V2X-Real dataset using CoBEVT as baseline. As shown in Figure \ref{fig:sup_vis_bev}, the observations are highly consistent with those on the simulated V2XSet dataset. In the clean setting, the broadcast paradigm exhibits significant inaccuracies within the blue circle due to semantic misalignment, while RSSA yields precise results by effectively correcting this misalignment. 

In the noisy setting, the broadcast paradigm remains inaccurate due to persistent semantic errors. Meanwhile, the handshake paradigm produces more false positives because it is extremely sensitive to pose noise during raw data alignment. In contrast, our RSSA maintains superior robustness and delivers the most accurate detection results, further confirming its effectiveness and reliability in complex real-world scenarios.

\section{Limitations}
\label{sec:limitations}
While our proposed RSSA demonstrates significant improvements in collaborative perception, there are a few limitations. Firstly, our validation is solely conducted on the 3D object detection. While RSSA's core contribution lies in robust feature alignment, its applicability and performance on other perception tasks, such as semantic segmentation or tracking, remain to be explored. Secondly, RSSA fundamentally relies on position information provided by the localization system and is not designed to operate under unknown pose conditions. Future work will aim to extend RSSA to a wider range of perception tasks and investigate its performance and adaptability in scenarios with completely unknown relative poses.

\newpage

\end{document}